# Predicting the Travel Distance of Patients to Access Healthcare using Deep Neural Networks

Li-Chin Chen, Ji-Tian Sheu, Yuh-Jue Chuang, Yu Tsao

***Abstract*—Objective:** Improving geographical access remains a key issue in determining the sufficiency of regional medical resources during health policy design. However, patient choices can be the result of the complex interactivity of various factors. The aim of this study is to propose a deep neural network approach to model the complex decision of patient choice in travel distance to access care, which is an important indicator for policymaking in allocating resources. **Method:** We used the 4-year nationwide insurance data of Taiwan and accumulated the possible features discussed in earlier literature. This study proposes the use of a convolutional neural network (CNN)-based framework to make predictions. The model performance was tested against other machine learning methods. The proposed framework was further interpreted using Integrated Gradients (IG) to analyze the feature weights. **Results:** We successfully demonstrated the effectiveness of using a CNN-based framework to predict the travel distance of patients, achieving an accuracy of 0.968, AUC of 0.969, sensitivity of 0.960, and specificity of 0.989. The CNN-based framework outperformed all other methods. In this research, the IG weights are potentially explainable; however, the relationship does not correspond to known indicators in public health, similar to common consensus. **Conclusions:** Our results demonstrate the feasibility of the deep learning-based travel distance prediction model. It has the potential to guide policymaking in resource allocation.

***Clinical and Translational Impact Statement*—**Deep learning technology is feasible in investigating the distance that patients would travel while accessing care. It is a tool that integrates complex interactive variables with highly imbalanced data distributions.

## I. INTRODUCTION

IT is the ideal for people to receive sufficient healthcare services without the need to travel a distance. However, nonmedical financial obstacles, such as transportation and high travel burdens, have been acknowledged as key barriers in accessing healthcare [1], [2]. Several studies have identified that patients are willing to travel further distances under certain circumstances. For example, [3] noted that patients with chronic illnesses travel approximately two-thirds of the distance to a physician compare to those who report no chronic illnesses. Further, [4] observed that half of the patients expressed that they would travel a distance to reduce the time on a waiting list for surgery. Other studies have identified an association between patient travel distance and disease severity [1], [4], [5]. However, studies have shown that the burden of travel makes treatment an unaffordable option [6]. Furthermore, [7] showed that patient health outcomes (for example, survival rates, length of hospital stay, and non-attendance at follow-up) decrease gradually as the distance to the healthcare facilities increases. Some studies noted that follow-up services must be geographically accessible to ensure utilization, irrespective of insurance status [8], [9].

Previous studies have used econometrics to predict patient choice. For example, [10] built patient choice models to predict the hospital that a patient would go to in the region using multinomial logit (MNL) and utility-maximizing nested logit. Furthermore, [11] distinguished patient choice between hospital-based and clinic-based care using a two-level nested MNL model, and [12] described the impact of quality on hospital choice using MNL. Generally, the purpose of econometrics is to estimate the interaction of

This research is funded by the Ministry of Science and Technology of Taiwan (MOST 109-2410-H-182-006 and MOST 110-2410-H-182-011-) and approved by the Research Ethics Committee at National Taiwan University (No. 202004EM035 and No. 202104EM038).

Li-Chin Chen is with the Research Center for Information Technology Innovation, Academia Sinica, 128 Academia Road, Section 2, Nankang, Taipei 115, Taiwan (li.chin@citi.sinica.edu.tw).

Ji-Tian Sheu is with the Department of Health Care Management, Chang Gung University, 259 Wen-Hwa 1st Road, Kwei-Shan, Taoyuan 333, Taiwan (jtsheu@mail.cgu.edu.tw).

Yuh-Jue Chuang is with the Department of Health Care Management, Chang Gung University, 259 Wen-Hwa 1st Road, Kwei-Shan, Taoyuan 333, Taiwan (chuangyj@mail.cgu.edu.tw).

Yu Tsao is with the Research Center for Information Technology Innovation, Academia Sinica, 128 Academia Road, Section 2, Nankang, Taipei 115, Taiwan (yu.tsao@citi.sinica.edu.tw).

The authors would like to thank Kai-Chun Liu, who is with the Research Center for Information Technology Innovation, Academia Sinica, for his friendly support.

variables and to explain causality, whereas it does not make predictions that indicate an action [10], [13].

Improving geographical access remains a key issue in health policy design [6], [17]. Travel distance is an essential piece of information that affects a patient's choice [10], [13]-[15]. Evaluating the travel distances of patients is a way of investigating whether the medical resources of the area are sufficient, estimating potential medical demands, and the tolerable distance that an individual is willing to travel [4], [16]. It provides important guidance for allocating resources from the perspective of policy [16], [18].

Most earlier studies mainly relied on conventional statistics and econometrics to analyze the impact of specific variables under certain hypotheses, preconditions, and limited patient groups. This involved in ruled out the potential confounding variables under balanced label sampling [13], [19]. However, it may be insufficient to support decision-making in reality when patient choices can result from a complex interaction of various factors and circumstances [4], [6], [19]-[22]. It is difficult to determine the insufficiency of regional medical resources without a tool that can integrate complex interactive variables and illustrate their interactivity under highly imbalanced circumstances. Such restrictions limit policymakers' decisions based on their experiences and interpretations.

Machine learning is known to be capable of processing multidimensional features and provides a generalized prediction [13], [23]. It has exhibited excellent performance in the medical domain, such as early risk detection [24], [25], mortality prediction [26], [27], symptom classification [28], and patient admission prediction [29]. The aim of this study is to propose a framework using a deep learning approach to predict the travel distance of a patient. The proposed approach has the potential to support decision making in policy design. To the best of our knowledge, this is the first study to demonstrate the feasibility of deep learning techniques in predicting the travel distance of patients to access healthcare.

The remainder of this paper is organized as follows. The methods section illustrates the retrieval of data, extraction of features, and the predicted target. Furthermore, it introduces machine learning methods, training strategies, evaluation indicators, and model interpretation methods. The results section illustrates the prediction and interpretation of the outcomes. The discussion section thoroughly interprets our findings. Finally, the conclusion section concludes our findings.

## II. METHODS AND PROCEDURES

The training and testing processes used in study are illustrated in Fig. 1. The following subsections introduce our method according to different phases, including data collection, data preprocessing, model training, prediction, performance evaluation, and interpretation of the prediction model.

### A. Data Collection

The data used in this research were the insurance claims from two million clinical declaration files and the Registry for Beneficiaries files from the Taiwan National Health Insurance Research Database (NHIRD), dated January 1, 2008 to December 31, 2011. The data released were originally sampled to ensure their representation of the population across Taiwan. The files included demographic information and visiting records of outpatients and emergency settings. In addition, four publicly announced data were included. One is the "physician density" information that referred to the number of practicing physicians serving per 10,000 people in each region of Taiwan [30]. Second, the national calendar was used to distinguish between workdays, weekends, and national holidays. Third, the center latitude and longitude of each district was used to calculate the travel distance of each visit. Fourth, the number of medical institutes and medical staff in each region, which were used to calculate the healthcare accessibility index (acc. index) based on the adjustment of the enhanced two-stage floating catchment area (2SFCA) method. The 2SFCA is a way to evaluate the local accessibility of medical care based on the regional physician-to-population ratio and the weight of distance decay effect (that is, the farther the distance is required to travel, the less likely an individual is to use a healthcare service) [2], [17], [31], [32].

This study was approved by the Research Ethics Committee at the National Taiwan University (No. 202004EM035 and No. 202104EM038) and waived the requirement for informed patient consent for the data, which had already been de-identified before analysis.

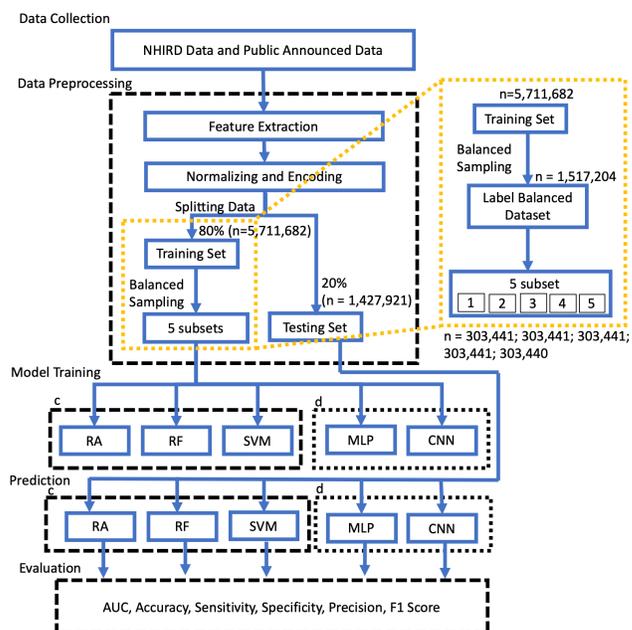

Fig. 1. Training and testing process flow. c indicates conventional machine learning methods, which underwent preprocessing of PCA and non-PCA. d indicates deep learning methods. RA: regression analysis; RF: random forest; SVM: support vector machine; MLP: multi-layer perceptron; CNN: convolutional neural network; PCA: principal component analysis; AUC:

area under the receiver operating characteristics curve. In the training dataset, the minimum number of traveled distances is 10–15 km (n = 379,301). The balanced label sampling used the under-sample strategy to randomly select data of other labels to reach an even amount of data, resulting in label balanced dataset having n = 1,517,204 (=379,301×4).

### B. Data Preprocessing

*Feature Extraction*

This study extracted 25 features that were accumulated from the earlier literature [34]-[39], as listed in Table I, and the detailed calculation equations are summarized in the Appendix [40]. Incomplete or questionable data, such as individuals without a birthdate or gender (or with two genders), records without a date, a birthdate later than the visit date, patients without any visiting records, patients without a primary diagnosis, incomplete information of visiting hospitals, patients unable to determine their place of residence (POR), and places that could not indicate the acc. index, were excluded. To avoid having dominated predictors, a Pearson correlation test was performed, presenting with a heatmap to demonstrate the correlation between all features and the prediction target.

TABLE I
FEATURES EXTRACTED FROM THE VISITING RECORDS

| Entity | Characteristics of the entity (Features) |
| --- | --- |
| Patients | Age, gender, low income (Yes/No), total number of visits, total number of diseases, total number of chronic diseases, usual provider of care (UPC), least usual provider of care (LUPC), sequential continuity of care index (SECOC), continuity of care index (COCI), and Charlson comorbidity index (CCI) score. |
| Providers | Practicing physicians serving per 10,000 people (physician density), the most frequent provider continuity (MFPC), the least frequent provider continuity (LFPC), region groups (north, central, south, east, and island) (Yes/No), and the healthcare accessibility index of each region (acc. index). |
| Incidents | Whether a surgery was involved (Yes/No), whether it was an emergency service (Yes/No), whether it was considered as a severe condition (Yes/No), whether the visit day was a workday (Yes/No), and the disease importance rate (DIR) of the target disease during that visit. |

*Definition of Travel Distance*

The targeted prediction outcome in this study was the travel distance of the patients. The POR and the location of the hospital were required to calculate the travel distance for each visit,. Owing to the nature of privacy protection, the NHIRD provides the POR of each patient with an approximate district that the patient registers as a hometown when filling the insurance form. However, registered districts often differ from where the patient actually lives. People may leave their hometown and live at an alternative location for several reasons, which makes it difficult to capture the travel distance. This study adopted the method proposed by Lin et al. [41] to obtain the estimated POR. It mainly used the records of treatment for flu, respiratory infection conditions, or emergency services to determine the POR of patients. These types of care services are less likely to travel far away. The estimated rules are shown in Fig. 2. This study identified the center of the POR district and the center of the hospital district with latitude and longitude, and calculated the distance between the two centers to obtain the estimated travel distance [2], [17], [31]. Considering the characteristics of the distance decay effect, impedance differentiations among areas [4], [32], [33], and that the distance was determined through approximation, we categorized the distance into four levels (< 5 km, 5–10 km, 10–15 km, and > 15 km). Approaching the prediction with approximate labels instead of actual values was considered to result in better generalization and more straightforward in practical use.

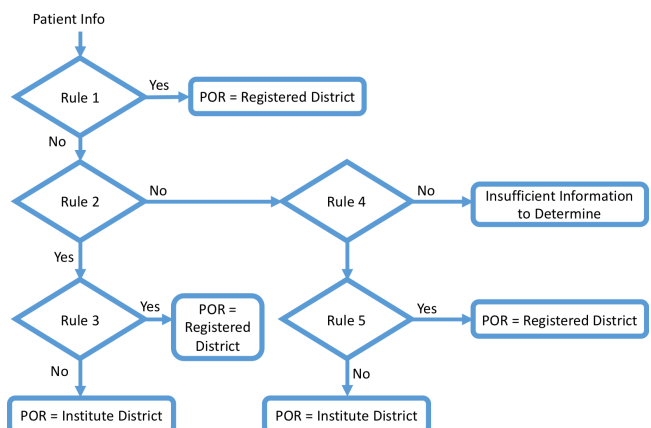

Fig. 2. Rules to determine the place of residence (POR): Adopting the queen contiguity definition, where all the districts with attached boundaries are included [42]. The nearby district list was generated using GeoDa version 1.14.0, an open geographic information system (GIS) software. The rules are slightly reduced because family members and relatives are not considered in this study.
Rule 1: Whether the patient belongs to a type A identity. Type A identity is the type of insurance identity that people are required to register at their POR.
Rule 2: Accessing care owing to flu or respiratory infection conditions.
Rule 3: Is the visiting institute in a nearby registered district?
Rule 4: At least having two emergency service records.
Rule 5: Is the emergency service institute in a nearby registered district?

*Data Normalization and Training Strategy*

After the preprocessing, all the numerical values (including age, number of diseases and chronic diseases, number of visits, number of votes as the most frequent provider continuity (MFPC), the least frequent provider continuity (LFPC), physician density, Charlson comorbidity index (CCI), accessibility index of each region, and disease importance rate (DIR)) were normalized between -1 and 1. The categorical features (including gender, regional groups, and the indication of low income, surgery, emergency service, severe condition, and workdays) were transformed into one-hot encoding [43], which converts a categorical variable that has $n$ values into $n$ variables. The numeric and categorical features were then concatenated into a patient visit vector to represent each event. Each physician visit was seemed as an independent visit event.

Afterward, the data were randomly split into training and testing data at an 80:20 ratio. Owning to the imbalanced distribution of patient travel distance, the training data were further randomly undersampled to reach a balanced label among the four distance labels. The training data were then randomly separated into five subsets. The training process included the rotation of each subset as a validation subset whereas the others acted as a training subset. On the other hand, considering that the models were required to face the actual behavior of patient travel, the testing data remained in its original label distribution, as the final evaluation of this research remains in its original label distribution.

*C. Model Training*

This study proposes a convolutional neural network (CNN)-based framework for travel distance prediction. The proposed framework was tested against three conventional machine learning methods and a deep learning method to demonstrate its effectiveness. The three conventional machine learning methods were regression analysis (RA), random forest (RF), and support vector machine (SVM). Because the conventional machine learning method does not inherit the ability of feature selection, principal component analysis (PCA) is added to conventional machine learning methods for feature reduction. That is, conventional machine learning methods underwent two preprocessing steps, one with PCA and the other without PCA. PCA is a multivariate statistical technique that extracts important information from the data and expresses it as a set of new orthogonal variables, known as the principal components [44], [45]. In our design, the number of components to be kept was set to 95%.

The deep learning method used was a multilayer perceptron (MLP) and the proposed CNN-based framework. Deep learning is known to inherit the ability to extract useful features automatically [46]-[48]; hence, PCA was not applied to MLP and CNN. The following are a brief introduction of the used methods used, including the conventional machine learning methods (i.e., RA, RF, and SVM) and deep learning methods (i.e., MLP and CNN):

*Regression Analysis*

Generally, RA is used to model and explain the relationship between a dependent variable y and other independent variables $x$. A standard model can be denoted as $y_i = x_i\beta + \varepsilon_i$, where $i = 1, ..., N$; $x\beta = x_1\beta_1 + \cdots + x_m\beta_m$. The estimated $\hat{\beta}$ is obtained by solving $X^T r = X^T(y - \hat{y})$, where $\hat{y} = X\hat{\beta}$ denotes the vector of the fitted value [49], [50]. In our design, the RA used the logistic regression method with a one-vs-rest scheme to model multiclass prediction.

*Random Forest*

RF is a tree-based classifier that ensembles the results of multiple decision trees. By setting the number of decision trees to be generated and the number of features to be selected, RF was tested for the best split when growing the trees. It returns the probabilities of the averaging classes of the produced trees for classification tasks or the general mean value of trees for regression tasks. The performance is considered better than that of a single classifier [51], [52]. In our design, the user-defined number of trees to be generated was set to 100, and the number of features to determine the best split was set to five.

*Support Vector Machine*

The basic idea of SVM is to establish a hyperplane that can maximize the distance between the plane and the nearest data [53], [54]. The hyperplane f(x) that separates the given data can be denoted as $f(x) = w^T x + b = \sum_{j=1}^{M} w_j x_j + b$, where M denotes the number of samples, and the inputs are $x_i$, where $i = \{1, 2, ..., M\}$. In our design, the C parameter, which indicates the tolerance degree of misclassification, was set to 0.2.

*Multi-Layer Perceptron*

MLP is a complex version of an artificial neural network that contains multiple hidden layers [55], [56], where every neuron in layer i is fully connected to every other neuron in layer $i + 1$. In a multi-layer neural network, each layer of the network is trained to produce a higher level of representation of the observed pattern [57], [58]. The computation of MLP can be denoted as $\hat{y} = \sigma\left(\sum_{j=1}^{d} x_j w_{ij} + b_{ij}\right)$, where each hidden layer computes a weight $w_{ij}$ and a bias $b_{ij}$ of the output from the previous layer, followed by a nonlinear activation function σ that calculates the sum as outputs. The number of units in the previous layer is represented by $d$, and the output of the previous layer is represented by $x_j$. In our design, the proposed MLP model contained 25 input nodes (based on input features) and five hidden layers containing 1500 neurons each. Rectified linear unit (ReLU) activation functions were used between each layer. Four output nodes symbolized the four categorized levels of distance.

*Convolutional Neural Network*

The concept of CNN is to extract meaningful information from the spatial pattern of data, which is primarily used for pattern recognition within images. CNNs comprise three types of layers: convolutional (Conv) layers, pooling layers, and fully-connected (FC) layers. The Conv layer uses a small array of numbers, called a kernel, to repeatedly apply across the input, and calculate an element-wise product between the input and the kernel to extract the spatial dimensionality, as shown in Fig. 3. The element-wise product is then summed up to obtain the output value in the corresponding position, called a feature map.

The pooling layers operate over each feature map; max-pooling, for example, takes only the maximum value of a certain size of matrix on the feature map, shown in Fig. 4, and denotes it as the feature in that section. Finally, the FC layer, which is analogous to the MLP form, predicts the final outcome based on the pooling results [59]-[61].

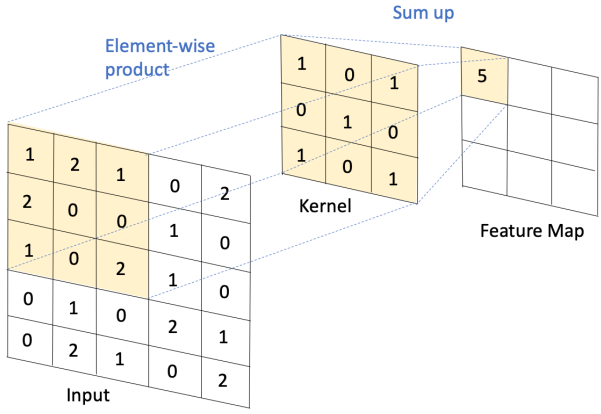

Fig. 3. An example of the convolution operation.

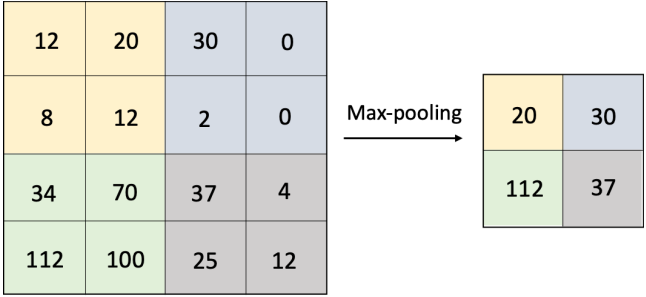

Fig. 4. An example of max-pooling operation.

In this study, we used a CNN to predict the distance that the patients would travel. The design of the proposed model is illustrated in Fig. 5. In our design, four one-dimensional convolutional (Conv1D) layers with a ReLU activation function and max-pooling were used. The convolution kernel size was set to 3, stride was set to 1, and the width of the max-pooling was 3. The FC layer had two hidden layers each containing 500 neurons. The final fully-connected output layer (FCO) consists of four output nodes to condense the result to a four-categorized output.

This study further demonstrated how adding layers to the proposed framework contributed to the prediction results and achieved optimization. The process included 1-layer Conv+1-layer FCO, 2-layer Conv+1-layer FCO, 3-layer Conv+1-layer FCO, 4-layer Conv+1-layer FCO, 4-layer Conv+1-layer FC+1-layer FCO, and finally, the proposed framework with 4-layer Conv+2-layer FC+1-layer FCO. A ReLU activation function was included in every FC layer to achieve nonlinear transformation.

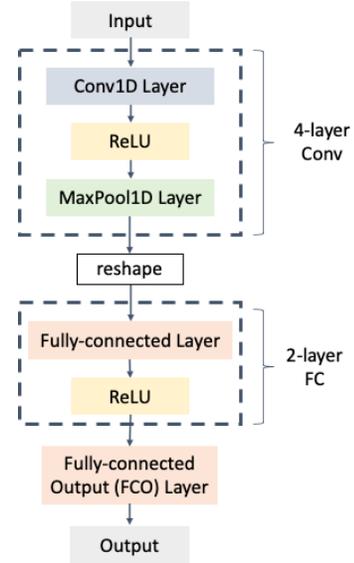

Fig. 5. Design of the proposed CNN-based framework. Conv1D: one-dimensional convolutional layer. ReLU: rectified linear unit. MaxPool1D: one-dimensional max-pooling layer. Conv: convolutional layer. FC: fully-connected layers.

### D. Model Evaluation

The prediction model was evaluated using indicators including the receiver operating characteristics (ROC) curve, area under the receiver operating characteristics curve (AUC), accuracy, sensitivity, specificity, precision, and F1 score [62]. (1)–(5) demonstrate the calculation of the indicators. For a multi-class classification of travel distance, the macro-average was used to generalize the performance index, which computed the metric independently for each class and then obtained the average to consider each class equally. The AUC used the one-vs-rest scheme to demonstrate the general performance.

$$\text{Accuracy} = (TP + TN) / (TP + FP + FN + TN) \quad (1)$$
$$\text{Sensitivity} = TP / (TP + FN) \quad (2)$$
$$\text{Specificity} = TN / (TN + FP) \quad (3)$$
$$\text{Precision} = TP / (TP + FP) \quad (4)$$
$$F1 = 2 \times (Prescision \times Sensitivity) / (Prescision + Sensitivity) \quad (5)$$

### E. Integrated Gradients Interpretation

Although deep learning models can achieve excellent performance, the model cannot explain how and why the model reaches its prediction, known as the black-box problem [47], [63]. The lack of transparency in the model is a serious barrier to implementing the application in practice [64]. This study explored the interpretation of the proposed model using the Integrated Gradients (IG) method [65]. Using a function $F: R^n \rightarrow [0,1]$ to represent a deep neuron network, let input $x \in R^n$ and baseline input $x' \in R^n$. IG considers the straight-line path from the baseline $x'$ to the input $x$, and computes all the integral gradients at all points along the path. The IG is obtained by cumulating these gradients, denoted as in (6), where the baseline is commonly chosen as near-zero $F(x') \approx 0$ so that it can be ignored and represents the weight of the individual input feature. $i$ represents the $i^{th}$ dimension along the $x$ and $x'$ paths.

$$IG_i(x) ::= (x_i - x'_i) \times \int_{\alpha=0}^{1} \frac{\partial F(x + \alpha \times (x - x'))}{\partial x_i} d\alpha \quad (6)$$

The IG was further averaged to obtain a weight for each input feature. We used an open-source interpretation library Captum [66] to implement the IG. This study was implemented with Python version 3.7.6, combined with PyTorch framework 1.1.0, scikit-learn 0.22.2, and Captum 0.3.1.

## III. RESULTS

A total of 7,139,603 visiting records of patients were included in the analysis, of which 65% of the patients had traveled less than 5 km, 17% had traveled 5–10 km, 6% had traveled 10–15 km, and 11% had traveled more than 15 km.

Demographic information of the patients is shown in Table II. The number of distance labels each patient traveled across accumulates up to 1.89 (SD = 0.80), and the headcount for each distance label shows that 92.45% of the patients have a record of choosing institutes that are nearby (< 5 km). Meanwhile, 32.01% of the patients had a record of choosing institutes that were far away (> 15 km).

According to the correlation heatmap shown in Fig. 6, six pairs of correlated features exceeded ±0.7: the total number of diseases and the total number of chronic diseases (0.998), MFPC and LFPC (0.986), age and CCI score (0.879), usual provider of care (UPC) and sequential continuity of care index (SECON) (0.778), the total number of visits and the total number of diseases (0.759), and the total number of visits and the total number of chronic diseases (0.758). None of the features appeared to be a dominat feature (correlation exceeding ±0.7) towards the distance, nor was the distance in level form or continuous value form.

The results indicated that the proposed CNN-based framework exceeded the performance of all other methods, achieving an accuracy of 0.968, AUC of 0.969, sensitivity of 0.960, and specificity of 0.989, as shown in Table V. Fig. 7 shows the ROC curve of the CNN-based framework against all other methods. Fig. 8 shows the contributions of the layers added to the proposed framework. The detailed performance values are listed Table VI in the appendix.

The interpretation results are shown in Fig. 9. The detailed values are listed Table VII in the appendix. The top three weighted features were the LFPC, physician density, and the number of chronic diseases. The last three weighted features were the MFPC, total number of visits, and SECOC.

TABLE II
DEMOGRAPHIC INFORMATION OF PATIENTS

| Item | | Value |
|---|---|---|
| Age, mean (SD) | | 49.331 (17.522) |
| Male, number of patients (%) | | 195,789 (46.819) |
| Noted low income, number of patients (%) | | 11,755 (2.811) |
| Total number of diseases, mean (SD) | | 17.758 (10.420) |
| Total number of chronic diseases, mean (SD) | | 17.607 (10.323) |
| Total number of visits per patient, mean (SD) | | 31.900 (24.111) |
| UPC, mean (SD) | | 0.480 (0.200) |
| LUPC, mean (SD) | | 0.085 (0.149) |
| COCI, mean (SD) | | 0.026 (0.061) |
| SECOC, mean (SD) | | 0.419 (0.217) |
| Number of distance labels per patient traveled across, mean (SD) | | 1.889 (0.798) |
| Number of patient headcounts | < 5 km, n (%) | 386,606 (92.450) |
| | 5–10 km, n (%) | 180,997 (43.282) |
| | 10–15 km, n (%) | 88,336 (21.124) |
| | > 15 km, n (%) | 133,868 (32.012) |

n = 418,180. n: number of patients; SD: standard deviation.

TABLE III
INFORMATION OF HOSPITALS

| Item | | Value |
|---|---|---|
| Hospital levels | Medical center, n (%) | 20 (0.103) |
| | Regional hospital, n (%) | 77 (0.398) |
| | District hospital, n (%) | 356 (1.840) |
| | Clinic, n (%) | 18,890 (97.658) |
| Region Groups | Northern, n (%) | 8,181 (42.294) |
| | Central, n (%) | 5,302 (27.410) |
| | Southern, n (%) | 5,364 (27.731) |
| | Eastern, n (%) | 446 (2.306) |
| | Island, n (%) | 50 (0.258) |
| MFPC, mean (SD) | | 45.205 (168.565) |
| LFPC, mean (SD) | | 45.085 (121.987) |
| Physician density, mean (SD) | | 24.410 (21.331) |
| Accessibility of the Region, mean (SD) | | 672,615.763 (641,445.063) |

n = 19,343. n: number of patients; SD: standard deviation.

TABLE IV
INFORMATION OF INCIDENTS

| Item | | Value |
|---|---|---|
| Hospital Levels | Medical center, n (%) | 584,048 (8.180) |
| | Regional hospital, n (%) | 747,859 (10.475) |
| | District hospital, n (%) | 573,836 (8.037) |
| | Clinic, n (%) | 5,233,860 (73.307) |
| Is surgery, n (%) | | 181,269 (2.539) |
| Is ER, n (%) | | 134,784 (1.888) |
| Is severe, n (%) | | 245,176 (3.434) |
| DIR, mean (SD) | | 0.1970 (0.183) |
| Workday, n (%) | | 5,963,772 (83.531) |
| Distance of travel (km), mean (SD) | | 9.012 (30.031) |
| Distance of travel (km) | < 5 km, n (%) | 4,667,188 (65.370) |
| | 5–10 km, n (%) | 1,229,860 (17.226) |
| | 10–15 km, n (%) | 474,010 (6.639) |
| | > 15 km, n (%) | 768,545 (10.765) |

n = 7,139,603. n: number of patients; SD: standard deviation.

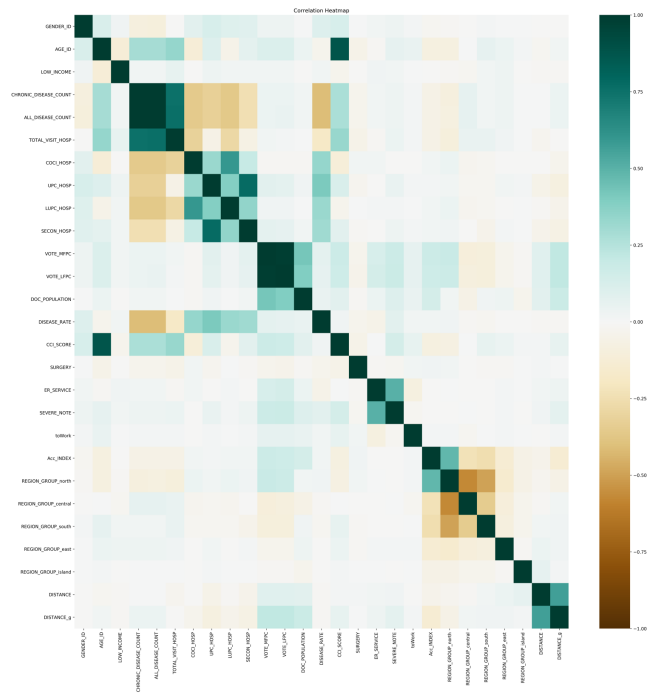

Fig. 6. Feature correlation heatmap. Correlation was calculated using the Pearson correlation test.

TABLE V
PREDICTION MODELS PERFORMANCE

|  | A | B | C | D | E | F | G | H* |
|---|---|---|---|---|---|---|---|---|
| AUC | 0.645 | 0.639 | 0.689 | 0.666 | 0.656 | 0.667 | 0.809 | 0.969 |
| Accuracy | 0.455 | 0.452 | 0.602 | 0.562 | 0.484 | 0.506 | 0.654 | 0.968 |
| F1 | 0.321 | 0.316 | 0.390 | 0.363 | 0.342 | 0.357 | 0.511 | 0.945 |
| Sensitivity | 0.362 | 0.356 | 0.403 | 0.384 | 0.376 | 0.391 | 0.556 | 0.960 |
| Specificity | 0.797 | 0.795 | 0.809 | 0.803 | 0.800 | 0.803 | 0.848 | 0.989 |
| Precision | 0.328 | 0.325 | 0.404 | 0.368 | 0.352 | 0.357 | 0.539 | 0.932 |

A: regression analysis without PCA; B: regression analysis with PCA; C: random forest without PCA; D: random forest with PCA; E: SVM without PCA; F: SVM with PCA; G: MLP; H*: CNN (proposed framework). PCA: principal component analysis.

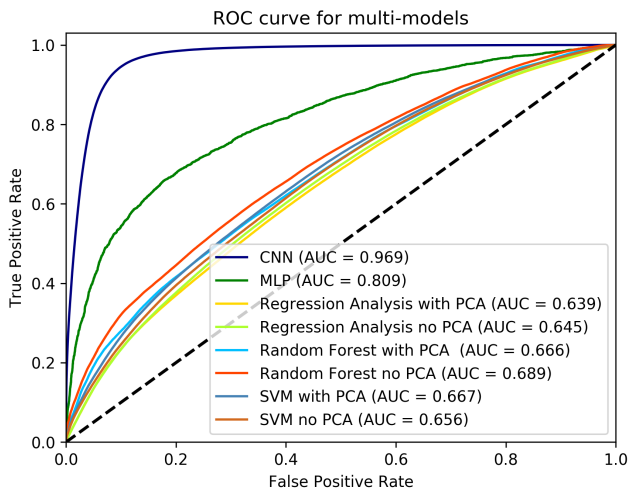

Fig. 7. Receiver operating characteristics (ROC) curve of the proposed model and all the other compared methods.

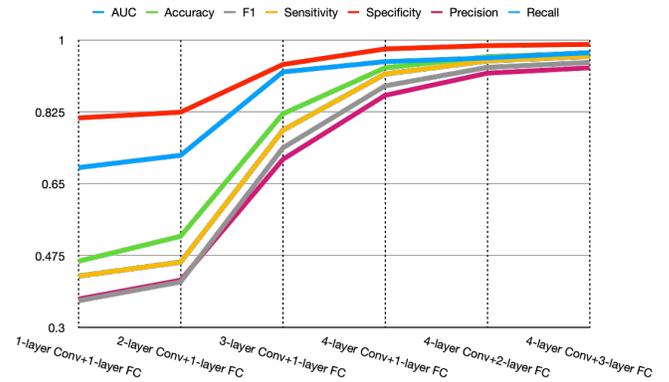

Fig. 8. Performance between the different layers of the proposed CNN-based framework.

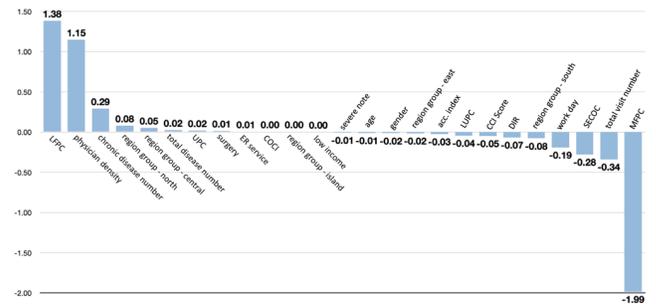

Fig. 9. Integrated Gradients feature interpretation of the CNN-based framework.

## IV. DISCUSSION

In our study, we successfully demonstrated the effectiveness of using a deep learning-based framework to predict the travel distance of patients. The results indicated that deep learning methods (MLP and CNN) outperformed conventional machine learning methods (RA, RF, and SVM) in processing structuralized insurance data. The proposed CNN-based framework outperformed all other methods. The performance converges and is optimized by adding layers by layers. Although the models were trained based on balanced label sampling, the results showed that the model was feasible when the testing data were highly imbalanced [19], [21]. This implies that the designed framework can be applied in real practice. In addition, we adopted a cross-validation strategy as an act of generalizing data distribution. The reported result is based on the validation of unseen data, which were isolated before the training process. This avoids the potential for overfitting in the prediction model. Meanwhile, based on the number of distance labels that a patient traveled and the number of headcounts for each distance label, both results indicate that patients may change their choice of travel distance under different circumstances; therefore, each visit event should be considered as independent.

Although conventional machine learning methods are known to require the process of feature selection, our results showed that, with or without the PCA process, performance did not differ much; nonetheless, the features used decreased from 25 to 16. This indicates that after PCA, conventional machine learning methods can use fewer features and resulting in an approximate performance. Notably, conventional machine learning methods were incapable of handling large amounts of data, whereas SVM failed to converge without increasing parameter c. The insurance and public health data include nationwide records, which are typically large and sparse. In contrast, the deep learning approach is known to extract useful information automatically without additional preprocessing [46]-[48]. This has the potential to achieve an ideal end-to-end system that requires no further interference after data input. In addition, it has no difficulty in processing a large volume of data. Combined with the observation of performance results, we conclude that deep neural networks are a better choice for implementation in public health problems such as predicting the travel distance of patients.

According to the IG interpretation results, the effect of the features on the prediction results can be positive or negative. Although all the features included were extracted based on earlier studies (meaning that they should all appear to be significant features), not all variables appear to be as effective. Some of the features are weighted approximately to zero, indicating that the scenario in Taiwan may appear differently. In general, patient preference (LFPC and MFPC), medical resources in each region (physician density and region groups), the complexity of diseases (number of chronic diseases), and the regularity of visits (total number of visits and SECOC) were weighted as the most effective features of the prediction results. It is worth noting that LFPC and MFPC are weighted as the top and last features, respectively, symbolizing that patient preference affects significantly and in different directions (positively and negatively). This indicates that the IG weights are potentially explainable and can be related to disciplinary knowledge in public health.

In addition, although some of the features shows correlated with each other in the Pearson correlation test, the calculated IG weight did not appear correspondingly (such as the total number of chronic diseases and the total number of diseases). Traditionally, acc. index and CCI score are considered more sophisticated indicators to identify the density of medical resources and the complexity of disease. Their IG weight did not exceed the value of physician density and the number of chronic diseases, which are indicators that provided less insight. Both results indicates that common consensus, such as correlation or sophisticated indicators, may not necessarily affect the prediction model simultaneously. The discipline of social science typically focuses on clarifying the causality and interrelationship of variables that affect patients' access to healthcare services. However, the machine learning approach attempted to provide a prediction that integrates the interaction results of the variables [79]. Further studies are required to interpret the differences in between.

Ensuring freedom of choice for patients is valued across countries. It has the potential to empower patients by prompting providers to compete for patients through a customer-market mechanism, improving care quality, efficiency, and wait time [67]-[76]. In Taiwan, freedom is assured under the universal coverage of National Health Insurance [77], [78], which is a perfect field for observing patient choice.

The decision-making process in policymaking generally involves current status investigation, policy design and evaluation, and post-implementation evaluation. Our work focused on supporting the first phase, whereas the model was capable of simulating patient choice with up to 96% accuracy. The policymaker can input patient data based on household registration in a particular region and investigate the distance that the patients would travel. Such information led to the evaluation of whether the patients considered the medical resources in the region to be sufficient or whether there are other considerations that induce them to travel further. In addition, the prediction result of patient choice has the potential to direct us in understanding patients' reactions under the current medical resource allocation. Based on the characteristics of patients' choices, the distance they would travel can be further analyzed, and the conclusion is informative in policy design. Our work led to a more precise current status investigation, and as a return, may potentially lead to more precise resource allocation.

The prediction was based on the trajectory of the de-identified patient-visit data commonly collected by insurance companies. Therefore, the model is highly achievable elsewhere, as it does not involve complex information that is difficult to collect or violates patient privacy. However, because of the nature of the de-identified data, the distance to travel can only be determined based on projection and assumption and cannot be validated accurately, which is a limitation of this study. Applications using deep learning technology are promising in healthcare policymaking, and further investigation is encouraged before industrialization, such as the discussion of individual impacts of each feature and its effect on the model performance and resource allocation.

## V. CONCLUSION

This study successfully demonstrated the effectiveness of using machine learning to predict the travel distance of patients. The proposed CNN-based framework performs well in processing structured insurance data. It was capable of handling complex combinations of features and imbalanced datasets, which is commonly noted when facing the problem of patient choice.

## APPENDIX

### A. Characteristics of the Patients

Age, gender, low income (Yes/No), total number of visits, total number of diseases, total number of chronic diseases, four continuity indicators, and a disease complexity index were included as characteristics of the patients. Age was determined according to the birthdate and visit date. The denotation of gender and low income was the status identified when entering the insurance. The total number of visits was calculated as the number of visiting records during the study period. The total number of diseases and chronic diseases were identified using the International Classification of Diseases (ICD) codes for each patient. The four continuity indicators captured the duration, density, dispersion, and continuity of an individual while accessing care [34]-[36]. To track the duration and density of patients accessing care, the UPC and least usual provider of care (LUPC) were used to indicate the visiting ratio of medical institutes. The UPC represented the most frequently visited institute, while LUPC represented the least frequently visited institute. The patients were required to visit the provider at least once to denote an institute as LUPC. The SECOC was used to calculate the change in providers, representing the dispersion of care. The continuity of care index (COCI) is a single indicator that represents the continuity of care for an individual. The calculation of the four continuity indicators is shown in (7) to (10), where N denotes the total number of visits, $n_i$ denotes the number of visits to the ith provider, k denotes the number of providers once visited, and $C_j$ is denoted as 1 when the jth provider is the same as the $(j+1)$th provider (0, if not). In addition, we included the CCI score [37] for the disease complexity of the individual.

$$\text{UPC} = \max\left(\frac{n_i}{N}\right) \quad (7)$$

$$\text{LUPC} = \min\left(\frac{n_i}{N}\right) \quad (8)$$

$$\text{SECOC} = \frac{\sum_{j=1}^{N-1} C_j}{N-1} \quad (9)$$

$$\text{COCI} = \frac{\left(\sum_{i=1}^{k} n_i^2\right) - N}{N(N-1)} \quad (10)$$

### B. Characteristics of the Providers

We characterized the providers using four indicators: the physician density of each region, MFPC, LFPC, and acc. index. Regarding patients' "vote with their feet" in choosing providers by themselves, we calculated the MFPC and LFPC to represent the patients' experiences and recommendations for each institute [38], [39]. The MFPC represents the frequency of being voted as the UPC, and the LFPC represents the frequency of being voted as the LUPC. Each patient was able to vote for only one MFPC or LFPC. The calculations are shown in (11) and (12), where p indicates the total number of patients, and $\text{UPC}_i$ and $\text{LUPC}_i$ indicate the ith patient who voted the provider as the *UPC* or *LUPC*, respectively. The acc. index of each region was calculated based on the adjustment of the enhanced two-stage floating catchment area method [2], [17], [31], [32], where the distance decay of each regional physician-to-population ratio is considered.

$$\text{MFPC} = \sum_{i=1}^{p} \text{UPC}_i \quad (10)$$

$$\text{LFPC} = \sum_{i=1}^{p} \text{LUPC}_i \quad (11)$$

### C. Characteristics of the Incidents

Each incident was characterized based on five encoded features. Through the encoded ICD codes and treatment codes of visiting records, we identified whether surgery was involved (Yes/No), whether it was an emergency service (Yes/No), whether it was considered as a severe condition (Yes/No), whether the visit day was a workday (Yes/No), and the disease importance rate (DIR) of the target disease during that visit. The identification of surgeries and emergency services was based on the treatment codes defined by the NHIRD. Severity was defined based on the emergency triage results. Triage results rank for levels 1 to 3 (1 = resuscitation, 2 = emergency, and 3 = urgent) were listed as severe, and the conditions that were included in the catastrophic illness announced by the National Health Insurance were also listed as severe. The date of the incident was categorized as workday or non-workday. We used the DIR to represent the importance of the target disease in that visit to determine whether the visit of the patient was a regular or singular event. The encoded primary diagnosis was identified as the target disease at that visit. The DIR represents the ratio of importance and is calculated as shown in (13), where N indicates the total number of visits to the patient, and $d_i$ indicates the total number of visits for disease $d_i$. We used only the primary diagnosis of the visit to identify the DIR.

$$\text{DIR} = \frac{d_i}{N} \quad (13)$$

TABLE VI
PERFORMANCE COMPARISON BETWEEN DIFFERENT LAYERS OF THE PROPOSED CNN-BASED FRAMEWORK

|  | 1-layer Conv+ 1-layer FCO | 2-layer Conv+ 1-layer FCO, | 3-layer Conv+ 1-layer FCO | 4-layer Conv+ 1-layer FCO | 4-layer CNN+ 2-layer FCN | 4-layer Conv+ 1-layer FC+1-layer FCO (Proposed Model) |
|---|---|---|---|---|---|---|
| AUC | 0.689 | 0.719 | 0.922 | 0.947 | 0.956 | 0.969 |
| Accuracy | 0.461 | 0.522 | 0.820 | 0.932 | 0.959 | 0.968 |
| F1 | 0.365 | 0.411 | 0.737 | 0.888 | 0.933 | 0.945 |
| Sensitivity | 0.425 | 0.459 | 0.780 | 0.917 | 0.949 | 0.960 |
| Specificity | 0.810 | 0.824 | 0.940 | 0.978 | 0.986 | 0.989 |
| Precision | 0.369 | 0.415 | 0.709 | 0.865 | 0.919 | 0.932 |
| Recall | 0.425 | 0.459 | 0.780 | 0.917 | 0.949 | 0.960 |

Conv: convolutional layer. FC: fully-connected layers. FCO: fully-connected output layer.

TABLE VII
INTEGRATED GRADIENTS FEATURE INTERPRETATION VALUES OF THE CNN-BASED FRAMEWORK

|  | Integrated Gradients Value |
|---|---|
| LFPC | 1.384 |
| physician density | 1.148 |
| chronic disease count | 0.293 |
| region group - north | 0.081 |
| region group - central | 0.051 |
| all disease count | 0.023 |
| UPC | 0.020 |
| surgery | 0.014 |
| ER service | 0.006 |
| COCI | 0.004 |
| region group - island | 0.004 |
| low income | 0.002 |
| severe note | -0.009 |
| age | -0.012 |
| gender | -0.015 |
| region group - east | -0.017 |
| acc index | -0.027 |
| LUPC | -0.045 |
| CCI | -0.050 |
| DIR | -0.069 |
| region group - south | -0.079 |
| workday | -0.191 |
| SECON | -0.279 |
| total visit count | -0.338 |
| MFPC | -1.985 |

LFPC: least frequent provider continuity. physician density: practicing physicians serving per 10,000 people. UPC: usual provider of care. COCI: continuity of care index. LUPC: least usual provider of care. CCI: Charlson comorbidity index. DIR: disease importance rate. SECOC: sequential continuity of care index. MFPC: most frequent provider continuity.

## REFERENCES

[1] N. N. Massarweh *et al.*, "Association between travel distance and metastatic disease at diagnosis among patients with colon cancer," *Journal of Clinical Oncology,* vol. 32, no. 9, p. 942, 2014.

[2] J. Saxon and D. Snow, "A rational agent model for the spatial accessibility of primary health care," *Annals of the American Association of Geographers,* vol. 110, no. 1, pp. 205-222, 2020.

[3] G. F. Nemet and A. J. Bailey, "Distance and health care utilization among the rural elderly," *Social Science & Medicine,* vol. 50, no. 9, pp. 1197-1208, 2000.

[4] M. Exworthy and S. Peckham, "Access, choice and travel: implications for health policy," *Social Policy & Administration,* vol. 40, no. 3, pp. 267-287, 2006.

[5] D. A. Axelrod *et al.*, "The interplay of socioeconomic status, distance to center, and interdonor service area travel on kidney transplant access and outcomes," *Clinical Journal of the American Society of Nephrology,* vol. 5, no. 12, pp. 2276-2288, 2010.

[6] V. Muralidhar, B. S. Rose, Y.-W. Chen, M. D. Nezolosky, and P. L. Nguyen, "Association between travel distance and choice of treatment for prostate cancer: does geography reduce patient choice?," *International Journal of Radiation Oncology* Biology* Physics,* vol. 96, no. 2, pp. 313-317, 2016.

[7] C. Kelly, C. Hulme, T. Farragher, and G. Clarke, "Are differences in travel time or distance to healthcare for adults in global north countries associated with an impact on health outcomes? A systematic review," *BMJ open,* vol. 6, no. 11, 2016.

[8] C. C. Lin *et al.*, "Association between geographic access to cancer care, insurance, and receipt of chemotherapy: geographic distribution of oncologists and travel distance," *Journal of Clinical Oncology,* vol. 33, no. 28, p. 3177, 2015.

[9] S. K. Schmitt, C. S. Phibbs, and J. D. Piette, "The influence of distance on utilization of outpatient mental health aftercare following inpatient substance abuse treatment," *Addictive Behaviors,* vol. 28, no. 6, pp. 1183-1192, 2003.

[10] H. Smith, C. Currie, P. Chaiwuttisak, and A. Kyprianou, "Patient choice modelling: how do patients choose their hospitals?," *Health care management science,* vol. 21, no. 2, pp. 259-268, 2018.

[11] D. E. Sahn, S. D. Younger, and G. Genicot, "The demand for health care services in rural Tanzania," *Oxford Bulletin of Economics and statistics,* vol. 65, no. 2, pp. 241-260, 2003.

[12] H. S. Luft *et al.*, "Does quality influence choice of hospital?," *Jama,* vol. 263, no. 21, pp. 2899-2906, 1990.

[13] S. Mullainathan and J. Spiess, "Machine learning: an applied econometric approach," *Journal of Economic Perspectives,* vol. 31, no. 2, pp. 87-106, 2017.

[14] T. C. Rosenthal and C. Fox, "Access to health care for the rural elderly," *JAMA,* vol. 284, no. 16, pp. 2034-2036, 2000.

[15] C. Y. Roh and M. J. Moon, "Nearby, but not wanted? The bypassing of rural hospitals and policy implications for rural health care systems," *Policy Studies Journal,* vol. 33, no. 3, pp. 377-394, 2005.

[16] A. M. Mosadeghrad, "Patient choice of a hospital: implications for health policy and management," *International journal of health care quality assurance,* 2014.

[17] M. R. McGrail, J. S. Humphreys, and B. Ward, "Accessing doctors at times of need–measuring the distance tolerance of rural residents for health-related travel," *BMC health services research,* vol. 15, no. 1, pp. 1-9, 2015.

[18] I. A. Glinos, R. Baeten, M. Helble, and H. Maarse, "A typology of cross-border patient mobility," *Health & place,* vol. 16, no. 6, pp. 1145-1155, 2010.

[19] Z. Obermeyer and E. J. Emanuel, "Predicting the future—big data, machine learning, and clinical medicine," *The New England journal of medicine,* vol. 375, no. 13, p. 1216, 2016.

[20] A. Victoor, D. M. Delnoij, R. D. Friele, and J. J. Rademakers, "Determinants of patient choice of healthcare providers: a scoping review," *BMC health services research,* vol. 12, no. 1, p. 272, 2012.

[21] J. Kleinberg, J. Ludwig, S. Mullainathan, and Z. Obermeyer, "Prediction policy problems," *American Economic Review,* vol. 105, no. 5, pp. 491-95, 2015.

[22] A. Victoor, D. M. Delnoij, R. D. Friele, and J. J. Rademakers, "Determinants of patient choice of healthcare providers: a scoping review," *BMC health services research,* vol. 12, no. 1, pp. 1-16, 2012.

[23] C. Petersen *et al.*, "Recommendations for the safe, effective use of adaptive CDS in the US healthcare system: an AMIA position paper," *Journal of the American Medical Informatics Association,* 2021.

[24] G. Alfian *et al.*, "Deep Neural Network for Predicting Diabetic Retinopathy from Risk Factors," *Mathematics,* vol. 8, no. 9, p. 1620, 2020.

[25] H. Zhou, R. Myrzashova, and R. Zheng, "Diabetes prediction model based on an enhanced deep neural network," *EURASIP Journal on Wireless Communications and Networking,* vol. 2020, no. 1, pp. 1-13, 2020.

[26] S. Raghunath *et al.*, "Prediction of mortality from 12-lead electrocardiogram voltage data using a deep neural network," *Nature medicine,* vol. 26, no. 6, pp. 886-891, 2020.

[27] K. Ikemura *et al.*, "Using Automated Machine Learning to Predict the Mortality of Patients With COVID-19: Prediction Model Development Study," *Journal of medical Internet research,* vol. 23, no. 2, p. e23458, 2021.

[28] U. R. Acharya *et al.*, "A deep convolutional neural network model to classify heartbeats," *Computers in biology and medicine,* vol. 89, pp. 389-396, 2017.

[29] R. El-Bouri, D. W. Eyre, P. Watkinson, T. Zhu, and D. A. Clifton, "Hospital admission location prediction via deep interpretable networks for the year-round improvement of emergency patient care," *IEEE Journal of Biomedical and Health Informatics,* vol. 25, no. 1, pp. 289-300, 2020.

[30] *Statistics Yearbook of Practicing Physicians and Health Care Organizations in Taiwan,* 2018. Available: https://www.tma.tw/stats/index_AllPDF.asp

[31] J. Bauer and D. A. Groneberg, "Measuring spatial accessibility of health care providers–introduction of a variable distance decay function within the floating catchment area (FCA) method," *PloS one,* vol. 11, no. 7, p. e0159148, 2016.

[32] W. Luo and Y. Qi, "An enhanced two-step floating catchment area (E2SFCA) method for measuring spatial accessibility to primary care physicians," *Health & place,* vol. 15, no. 4, pp. 1100-1107, 2009.

[33] J. E. Sherman, J. Spencer, J. S. Preisser, W. M. Gesler, and T. A. Arcury, "A suite of methods for representing activity space in a healthcare accessibility study," *International journal of health geographics,* vol. 4, no. 1, pp. 1-21, 2005.

[34] C. E. Pollack, P. S. Hussey, R. S. Rudin, D. S. Fox, J. Lai, and E. C. Schneider, "Measuring care continuity: a comparison of claims-based methods," *Medical care,* vol. 54, no. 5, pp. e30-e34, 2016.

[35] C.-L. Chan, H.-J. You, H.-T. Huang, and H.-W. Ting, "Using an integrated COC index and multilevel measurements to verify the care outcome of patients with multiple chronic conditions," *BMC health services research,* vol. 12, no. 1, p. 405, 2012.

[36] MCHP. "Concept: Measuring Majority of Care." Manitoba Centre for Health Policy. http://mchp-appserv.cpe.umanitoba.ca/viewConcept.php?conceptID=1108 (accessed 2019.04.03).

[37] M. E. Charlson, P. Pompei, K. L. Ales, and C. R. MacKenzie, "A new method of classifying prognostic comorbidity in longitudinal studies:

development and validation," *Journal of chronic diseases,* vol. 40, no. 5, pp. 373-383, 1987.

[38] A. Higgins, T. Zeddies, and S. D. Pearson, "Measuring the performance of individual physicians by collecting data from multiple health plans: the results of a two-state test," *Health affairs,* vol. 30, no. 4, pp. 673-681, 2011.

[39] C. Saint-Pierre, F. Prieto, V. Herskovic, and M. Sepúlveda, "Relationship between Continuity of Care in the Multidisciplinary Treatment of Patients with Diabetes and Their Clinical Results," *Applied Sciences,* vol. 9, no. 2, p. 268, 2019.

[40] L. Chen, Y. Tsao, and J.-T. Sheu, "Using Deep Learning and Explainable Artificial Intelligence in Patients' Choices of Hospital Levels," *arXiv preprint arXiv:2006.13427,* 2020.

[41] M.-H. Lin, A.-C. Yang, and T.-H. Wen, "Using Regional Differences and Demographic Characteristics to Evaluate the Principles of Estimation of the Residence of the Population in National Health Insurance Research Databases (NHIRD)," *Taiwan Journal of Public Health,* vol. 30, no. 4, pp. 347-361, 2011.

[42] L. Anselin. "Contiguity-Based Spatial Weights." Available: https://geodacenter.github.io/workbook/4a_contig_weights/lab4a.html#queen-contiguity

[43] P. Rodríguez, M. A. Bautista, J. Gonzalez, and S. Escalera, "Beyond one-hot encoding: Lower dimensional target embedding," *Image and Vision Computing,* vol. 75, pp. 21-31, 2018.

[44] S. Wold, K. Esbensen, and P. Geladi, "Principal component analysis," *Chemometrics and intelligent laboratory systems,* vol. 2, no. 1-3, pp. 37-52, 1987.

[45] R. Bro and A. K. Smilde, "Principal component analysis," *Analytical methods,* vol. 6, no. 9, pp. 2812-2831, 2014.

[46] B. Shickel, P. J. Tighe, A. Bihorac, and P. Rashidi, "Deep EHR: a survey of recent advances in deep learning techniques for electronic health record (EHR) analysis," *IEEE journal of biomedical and health informatics,* vol. 22, no. 5, pp. 1589-1604, 2018.

[47] C. Xiao, E. Choi, and J. Sun, "Opportunities and challenges in developing deep learning models using electronic health records data: a systematic review," *Journal of the American Medical Informatics Association,* vol. 25, no. 10, pp. 1419-1428, 2018.

[48] R. Miotto, L. Li, B. A. Kidd, and J. T. Dudley, "Deep patient: an unsupervised representation to predict the future of patients from the electronic health records," *Scientific reports,* vol. 6, no. 26094, 2016.

[49] D. Pregibon, "Logistic regression diagnostics," *Annals of statistics,* vol. 9, no. 4, pp. 705-724, 1981.

[50] C.-Y. J. Peng, K. L. Lee, and G. M. Ingersoll, "An introduction to logistic regression analysis and reporting," *The journal of educational research,* vol. 96, no. 1, pp. 3-14, 2002.

[51] M. Belgiu and L. Drăguţ, "Random forest in remote sensing: A review of applications and future directions," *ISPRS journal of photogrammetry and remote sensing,* vol. 114, pp. 24-31, 2016.

[52] T. M. Oshiro, P. S. Perez, and J. A. Baranauskas, "How many trees in a random forest?," in *International workshop on machine learning and data mining in pattern recognition,* 2012: Springer, pp. 154-168.

[53] W. S. Noble, "What is a support vector machine?," *Nature biotechnology,* vol. 24, no. 12, pp. 1565-1567, 2006.

[54] A. Widodo and B.-S. Yang, "Support vector machine in machine condition monitoring and fault diagnosis," *Mechanical systems and signal processing,* vol. 21, no. 6, pp. 2560-2574, 2007.

[55] W. Liu, Z. Wang, X. Liu, N. Zeng, Y. Liu, and F. E. Alsaadi, "A survey of deep neural network architectures and their applications," *Neurocomputing,* vol. 234, pp. 11-26, 2017.

[56] V. Sze, Y.-H. Chen, T.-J. Yang, and J. S. Emer, "Efficient processing of deep neural networks: A tutorial and survey," *Proceedings of the IEEE,* vol. 105, no. 12, pp. 2295-2329, 2017.

[57] R. Miotto, F. Wang, S. Wang, X. Jiang, and J. T. Dudley, "Deep learning for healthcare: review, opportunities and challenges," *Briefings in bioinformatics,* vol. 19, no. 6, pp. 1236-1246, 2018.

[58] A. M. Bur, M. Shew, and J. New, "Artificial intelligence for the otolaryngologist: a state of the art review," *Otolaryngology–Head and Neck Surgery,* vol. 160, no. 4, pp. 603-611, 2019.

[59] K. O'Shea and R. Nash, "An introduction to convolutional neural networks," *arXiv preprint arXiv:1511.08458,* 2015.

[60] R. Yamashita, M. Nishio, R. K. G. Do, and K. Togashi, "Convolutional neural networks: an overview and application in radiology," *Insights into imaging,* vol. 9, no. 4, pp. 611-629, 2018.

[61] A. Zaitcev, M. R. Eissa, Z. Hui, T. Good, J. Elliott, and M. Benaissa, "A Deep Neural Network Application for Improved Prediction of HbA1c in Type 1 Diabetes," *IEEE journal of biomedical and health informatics,* vol. 24, no. 10, pp. 2932-2941, 2020.

[62] R. Parikh, A. Mathai, S. Parikh, G. C. Sekhar, and R. Thomas, "Understanding and using sensitivity, specificity and predictive values," *Indian journal of ophthalmology,* vol. 56, no. 1, p. 45, 2008.

[63] M. V. García and J. L. Aznarte, "Shapley additive explanations for NO2 forecasting," *Ecological Informatics,* vol. 56, p. 101039, 2020.

[64] G. Montavon, S. Lapuschkin, A. Binder, W. Samek, and K.-R. Müller, "Explaining nonlinear classification decisions with deep taylor decomposition," *Pattern Recognition,* vol. 65, pp. 211-222, 2017.

[65] M. Sundararajan, A. Taly, and Q. Yan, "Axiomatic attribution for deep networks," in *International Conference on Machine Learning,* 2017: PMLR, pp. 3319-3328.

[66] N. Kokhlikyan *et al.*, "Captum: A unified and generic model interpretability library for pytorch," *arXiv preprint arXiv:2009.07896,* 2020.

[67] A. Victoor, D. M. Delnoij, R. D. Friele, and J. J. Rademakers, "Determinants of patient choice of healthcare providers: a scoping review," *BMC health services research,* vol. 12, p. 272, 2012.

[68] A. Victoor, R. D. Friele, D. M. Delnoij, and J. J. Rademakers, "Free choice of healthcare providers in the Netherlands is both a goal in itself and a precondition: modelling the policy assumptions underlying the promotion of patient choice through documentary analysis and interviews," *BMC health services research,* vol. 12, p. 441, 2012.

[69] N. Reibling and C. Wendt, "Gatekeeping and provider choice in OECD healthcare systems," *Current Sociology,* vol. 60, no. 4, pp. 489-505, 2012.

[70] M. Kaneko *et al.*, "Gatekeeping function of primary care physicians under Japan's free-access system: a prospective open cohort study involving 14 isolated islands," *Family practice,* vol. 36, no. 4, pp. 452-459, 2019.

[71] K. Hoffmann, A. George, T. Van Loenen, J. De Maeseneer, and M. Maier, "The influence of general practitioners on access points to health care in a system without gatekeeping: a cross-sectional study in the context of the QUALICOPC project in Austria," *Croatian medical journal,* vol. 60, no. 4, p. 316, 2019.

[72] J. Kaipio *et al.*, "Usability problems do not heal by themselves: National survey on physicians' experiences with EHRs in Finland," *International Journal of Medical Informatics,* vol. 97, pp. 266-281, 2017.

[73] D. King *et al.*, "Identifying quality indicators used by patients to choose secondary health care providers: a mixed methods approach," *JMIR mHealth and uHealth,* vol. 3, no. 2, p. e65, 2015.

[74] M. Dumontet, T. Buchmueller, P. Dourgnon, F. Jusot, and J. Wittwer, "Gatekeeping and the utilization of physician services in France: Evidence on the Médecin traitant reform," *Health Policy,* vol. 121, no. 6, pp. 675-682, 2017.

[75] H. S. Suh, H.-Y. Kang, J. Kim, and E. Shin, "Effect of health insurance type on health care utilization in patients with hypertension: a national health insurance database study in Korea," *BMC health services research,* vol. 14, no. 570, p. 570, 2014.

[76] M. Kaneko and M. Matsushima, "Current trends in Japanese health care: establishing a system for board-certificated GPs," *British Journal of General Practice,* vol. 67, no. 654, pp. 29-29, 2017.

[77] J.-F. R. Lu and W. C. Hsiao, "Does universal health insurance make health care unaffordable? Lessons from Taiwan," *Health affairs,* vol. 22, no. 3, pp. 77-88, 2003.

[78] T.-Y. Wu, A. Majeed, and K. N. Kuo, "An overview of the healthcare system in Taiwan," *London journal of primary care,* vol. 3, no. 2, pp. 115-119, 2010.

[79] S. M. Lundberg *et al.*, "Explainable machine-learning predictions for the prevention of hypoxaemia during surgery," *Nature biomedical engineering,* vol. 2, no. 10, pp. 749-760, 2018.